\UseRawInputEncoding
\documentclass[]{IEEEtran}
\usepackage[inner=0.9055in,outer=0.5118in,top=0.5in]{geometry}
\pdfoutput=1
\usepackage{graphicx}
\usepackage{amsmath}
\usepackage{amssymb}			

\usepackage{enumitem}			

\usepackage{amsthm}								

\usepackage[numbers, sort & compress]{natbib} 	

\begin{document}
\title{Causal reasoning in typical computer vision tasks}

\author{Kexuan Zhang$^\S$, Qiyu Sun$^\S$, Chaoqiang Zhao$^\S$, Yang Tang$^\S{}^{*}$\\$^\S$Key Laboratory of Advanced Control and Optimization for Chemical Process, Ministry of Education,\\East China University of Science and Technology, Shanghai 200237, China\\}
\maketitle

\begin{abstract}
Deep learning has revolutionized the field of artificial intelligence. Based on the statistical correlations uncovered by deep learning-based methods, computer vision has contributed to tremendous growth in areas like autonomous driving and robotics. Despite being the basis of deep learning, such correlation is not stable and is susceptible to uncontrolled factors. In the absence of the guidance of prior knowledge, statistical correlations can easily turn into spurious correlations and cause confounders. As a result, researchers are now trying to enhance deep learning methods with causal theory. Causal theory models the intrinsic causal structure unaffected by data bias and is effective in avoiding spurious correlations. This paper aims to comprehensively review the existing causal methods in typical vision and vision-language tasks such as semantic segmentation, object detection, and image captioning. The advantages of causality and the approaches for building causal paradigms will be summarized. Future roadmaps are also proposed, including facilitating the development of causal theory and its application in other complex scenes and systems.
\end{abstract}


\section{Introduction}\label{section 1}
Deep learning techniques have improved our understanding of the world, and deep learning-based computer vision methods enable high-performance intelligent perception of our surroundings \cite{scideep}. Areas such as autonomous cars \cite{scicar,li2019challenges}, unmanned aircraft \cite{techau}, and robotics \cite{scirobotics,scirob} are developing rapidly with technological innovations. To guarantee performance across one or even more domains, training strategies like attention mechanisms \cite{attention,attentionsci}, pre-training mechanisms \cite{pre-train,pretrain}, and generic large models \cite{li2023blip} have been proposed. Despite their great performance, the basis of these deep learning-based methods is to learn the statistical correlation. However, statistical correlation learns regular knowledge based on the final presentation of the data, which lacks the guidance of prior knowledge. In contrast, causality focuses on the mechanisms of the data generation process and the causal structure of specific tasks.
Causality implies the relationship between two variables in 
which the cause variable directly affects the effect variable \cite{pearl2009causality}. 
Though both statistical correlation and causality are data-driven \cite{gao2022causal}, the former uses the consistency of data trends as the basis for determining relationships, while causality is the reflection of inherent characteristics and structures both inside and between variables. As there is not always a causal relationship between variables that change in a consistent trend, statistical correlation-based methods tend to use mistakenly non-causal relationships, also called spurious correlations, as the basis for network inference. In particular, when the data is multidimensional and heterogeneous, the complexity of the relationships may further amplify the impact of spurious correlations. Simpson's paradox proposed by Blyth \emph{et al.} \cite{simpson} well confirms the flaws of statistical correlation-based methods. It confirms that the direction of an association at the population-level may be reversed within the subgroups comprising that population. As stated in \cite{Borsboom2009}, the relation between coffee consumption and neuroticism is positive in each individual, but those individuals who drink more coffee are generally less neurotic. For each individual, the correlation between coffee consumption and neuroticism is positive, but in the population the correlation is negative. Such paradox is worth noticing as different levels of interpretation lead to different results for the same data. It follows that neither individual-level nor population-level statistical correlations can fully characterize the relationship between coffee consumption and neuroticism. However, the causality-based methods will specify the prior knowledge concerning the causal structures and derive the correct causal chains at the specific level of interpretation. As a result, the causality-based methods are more logical and effective compared to the statistical correlation-based ones. \\
Typical deep learning-based computer vision tasks can be summarized as follows: Given an image $X$, the goal is to build a network to predict its label $Y$ correctly \cite{malik2019deep}. A statistical model fitted with a suitable objective function is often used to estimate the conditional probability distribution $P(Y|X)$. However, only under the independent identical distribution ($\mathcal{I.I.D.}$) hypothesis can the learned conditional probability distribution $P (Y|X)$ be applied appropriately from the training set to the testing set. It requires that new prediction samples stay consistent with the distribution of the training set. To minimize the impact of domain differences between two sets, deep learning-based methods such as domain adaptation \cite{sun2022survey} and generalization \cite{zhou2021domain} have been proposed. However, the problem of generalizability cannot be fundamentally solved through these methods, as the bias caused by domain differences is not fundamentally eliminated. Additionally, since deep learning-based methods benefit from many stacked layers and parameters for approximating high-dimensional complex functions, it is hard to interpret the learned models and parameters \cite{malik2019deep}. In conclusion, existing statistical correlation-based methods overly rely on the given data and analyze the problems through approximating high-dimensional functions rather than mechanical mining. There are areas for improvement in both generalizability and interpretability.
\begin{figure*}[t]
    \centering
    \includegraphics[height=9cm,width=16cm]{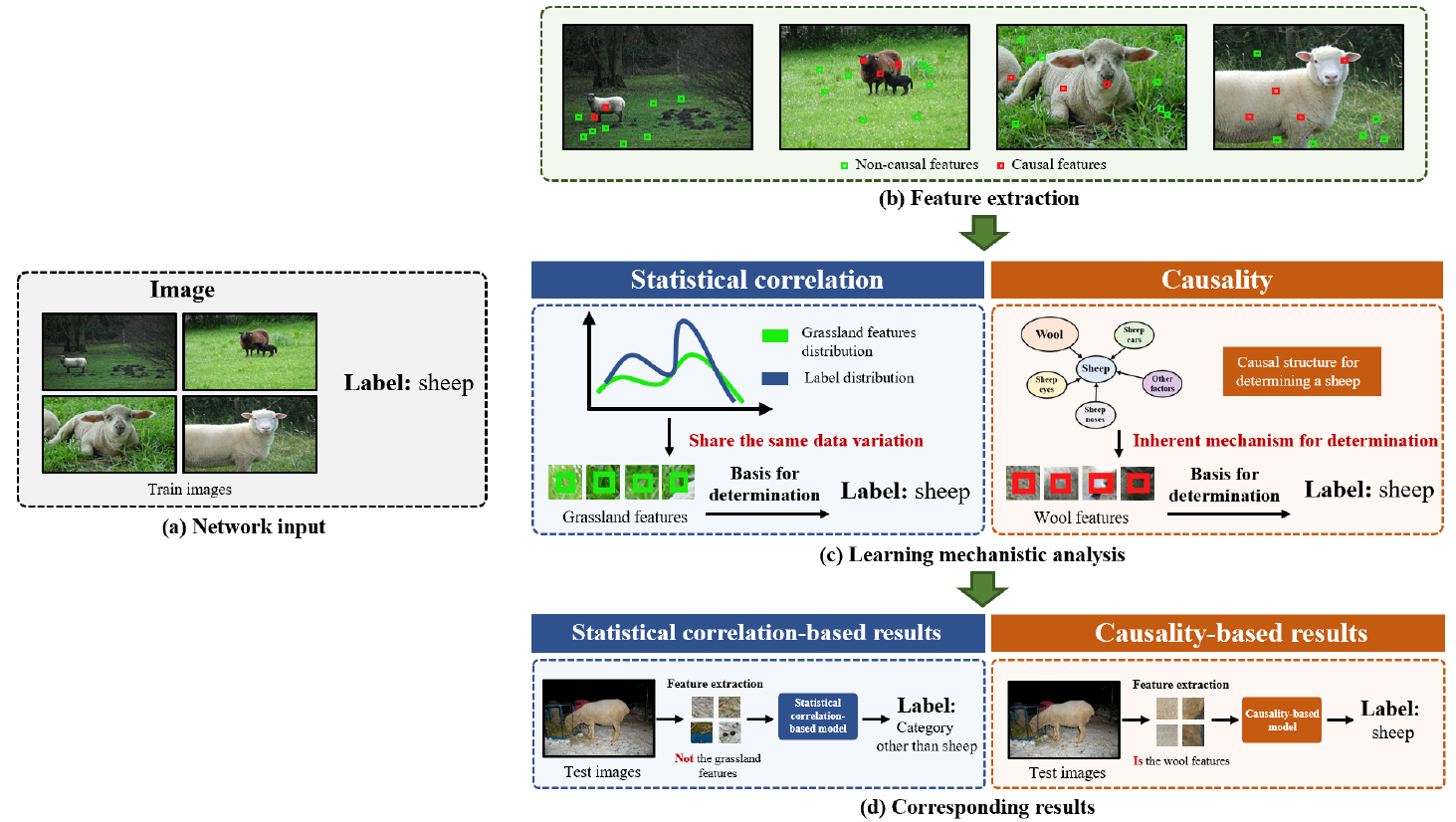}
    \caption{Comparison between statistical correlation-based methods and causality-based methods. Given the images and the corresponding labels, the image classification network performs feature extraction, makes an analytical decision on the feature, and finally outputs the corresponding labels. The figure above compares the two learning mechanisms: statistical correlation-based analysis and causality-based analysis, as shown in (c). The input in (a) is visualized by feature extraction in (b). The green boxes indicate the captured non-causal features, while the red boxes indicate the captured causal features. Due to different learning mechanisms, the statistical correlation-based method fails to infer the correct label. In contrast, the causality-based method can get the correct label, as shown in (d). When faced with the learned object in an unknown environment, the correlation-based method tends to be misled by data bias. Contrastly, the causality-based method focuses only on the causal factors associated with the object and is not disturbed by data variation.}
    \label{ex}
\end{figure*}\\
Due to its mechanical strengths, causality has gained significant attention recently and has rooted its developments across several fields, such as statistics \cite{heidel2016causality,dawid2015statistical}, economics \cite{heckman2022causality,geweke1984inference}, epidemiology \cite{kundi2006causality,ohlsson2020applying}, and computer science \cite{hair2021data,prosperi2020causal}. Basic causal methods can be divided into two main aspects, causal discovery and causal inference \cite{chen2022review}. Based on the causal structures learned by causal discovery, causal inference leverages those relationships for further analysis. An example is proposed to highlight the advantages of causality-based methods. Fig. \ref{ex} compares the statistical correlation-based and the causality-based methods for the same image classification task. Given the input images of the sheep and the corresponding labels in Fig. \ref{ex} (a), the model is trained for correct identification. The visualizations of the learned features are shown in Fig. \ref{ex} (b). Both statistical correlation-based and causality-based learning mechanisms are specifically shown in Fig. \ref{ex} (c). Since the sheep and the grassland frequently coexist in the training data, the statistical correlation-based process tends to regard the grassland characteristics as the basis for the labels due to their similar distribution. In contrast, the causality-based methods focus more on the objective cause chain of sheep and prefer wool features as the basis. Different learning mechanisms lead to different classification results, especially when faced with rare samples, as shown in Fig. \ref{ex} (d). When given an image of a sheep standing in the snow, the statistical correlation-based model fails to assign the correct label because no grassland features are found. Conversely, by focusing on the wool features, the causality-based model can accurately categorize the sheep based on the causal features. As a result, the causality-based methods rely not only on the consistency of data trends but also on the underlying processes that generate the causal structure between variables \cite{pearl2011bayesian}. They are more robust towards an unknown environment, and the learning process tends to be more interpretable.\\
This review focuses on the implications of causal theory for vision and vision-language tasks, including classification, detection, segmentation, visual recognition, image captioning, and visual question answering. There exist some surveys about causal theory \cite{kaddour2022causal, gao2022causal, li2023survey, chen2022review}. Kaddour \emph{et al.} \cite{kaddour2022causal} group existing causal machine learning work into five categories and comprehensively compare existing approaches. Gao \emph{et al.} \cite{gao2022causal} focus on the application of causal reasoning to recommendation systems. Li \emph{et al.} \cite{li2023survey} outline the advantages of causal theory in industrial applications. Chen \emph{et al.} \cite{chen2022review} divide the causal discovery tasks into three types according to the variable paradigm: definite tasks, semidefinite tasks, and undefinite tasks. Unlike these existing causality-based reviews, this work compares the causal structures of different approaches in vision and vision-language tasks and summarizes the correspondence between the commonly used causal structures and the corresponding concerns. Furthermore, the article outlines the benefits of the causal approaches regarding accuracy, generalizability, and interpretability. Section \ref{section 2} introduces the basic concepts and terminology of causality, including structural causal models, causal interventions, back-door adjustments, front-door adjustments, and counterfactual. Section \ref{section 3} proposes a systematic and structural review of the specific vision and vision-language tasks. It analyzes the corresponding causal structure and identifies the confounders. Five typical causal structures are summarized for comparison and methodology overviews. Section \ref{section 4} lists the future roadmaps for better development of causal theory and a broad application of causal theory. Section \ref{section 5} concludes this review. 
\section{Preliminaries}\label{section 2}
In this section, the basic causal methods are introduced, including causal discovery and causal inference. Causal discovery aims to build causal relationships with the structural causal models (SCMs), while causal inference is used to estimate the causal effect.\\
\subsection{Causal discovery}\label{section 2.2}
\subsubsection{Causal structure}
\begin{figure*}[t]
    \centering
    \includegraphics[height=3cm, width=6cm]{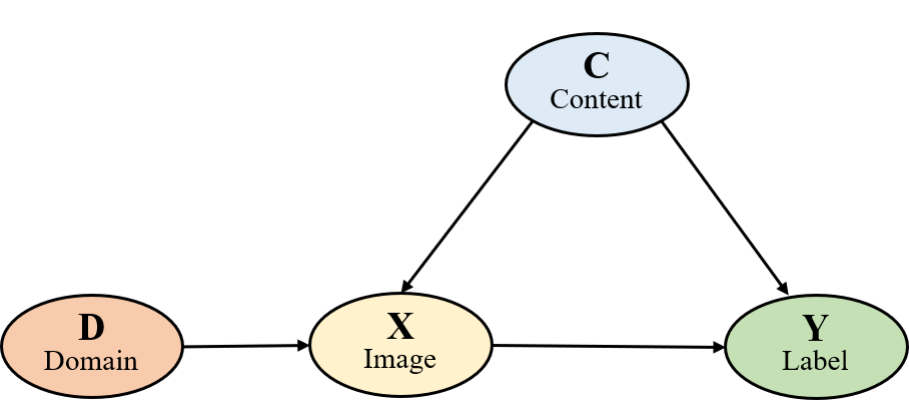}
    \caption{A general causal structure for image generation, where $X$ represents the given image, $Y$ represents the learned label, $C$ represents the corresponding content, and $D$ represents the domain involved.}
    \label{data process}
\end{figure*}

A causal structure is a directed acyclic graph (DAG) in which each node corresponds to a specific variable, and each edge indicates a direct functional relationship between the linked variables \cite{pearl2009causality}. If there is a directed edge pointing from $Y$ to $X$, $X$ is called the child variable, while $Y$ is the corresponding parent variable. If a variable has its parent variable in the causal structure, it is endogenous; otherwise, it is exogenous. To better describe the concepts involved, the causal structure depicting the process of image generation is shown in Fig. \ref{data process} as an example. \\
The graph has four nodes involved, where $X$ represents the given image, $Y$ represents the corresponding label, $C$ represents the content, and $D$ represents the domain involved. The edge $D \xrightarrow{} X \xleftarrow{} C$ describes image generation process in that the image includes both the content and the domain information. The edge $X \xrightarrow{} Y$ indicates the ultimate goal of a vision task: finding a suitable label for the given image. The edge $C \xrightarrow{} Y$ means that the content in the image determines the label. Since there are two directed edges pointing to X: $D \xrightarrow{} X$ and $C \xrightarrow{} X$, D and C are the parent variables of $X$. Also, $D$ is exogenous because it has no parent variables in the graph. A path between two variables is a sequence of edges connecting them. The path between $X$ and $Y$ in Fig. \ref{data process} can be either $X \xrightarrow{} Y$ or $X \xleftarrow{} C \xrightarrow{} Y$.\\
\subsubsection{Structural causal models}\label{section 2.2.1}
Based on the causal structure, a structural causal model can specify how each variable is influenced by its parent variables by effect functions. Given a set of variables $X_1, X_2...,X_n$ indicating the nodes in the causal structure, they can be written as the outcome of their parent variables and the effect function \cite{pearl2009causality}.
\begin{equation}
    X_i = f_i(PA_i,U_i), i=1,...,n,
\end{equation}
where $PA_i$ indicates the parent variables of $X_i$, $U_i$ are the unobserved background variables including the noise, and $f_i$ indicates the functional relationship. Taking the relationship of variables in Fig. \ref{data process} as an example, the variable $X$ is the outcome of its parent variables and the effect functions together as follows:
\begin{equation}
    X = f(D,C,U),
\end{equation}
where $f$ is the effect function, $U$ indicates the background variable in this functional relationship. The purpose of the causal inference to be presented later is to characterize $f$ and estimate the effects of interventions or counterfactuals.
\subsection{Causal inference}
Given a causal structure, causal intervention and counterfactuals can be conducted. The former aims to change the structure by specifying variable data to artificially alter the relationship between variables. In contrast, the latter aims to estimate the variables in a situation opposite to reality. Two typical intervention methods: back-door adjustment and front-door adjustment will also be discussed.

\subsubsection{Causal intervention}
\begin{figure*}[t]
    \centering
    \includegraphics[height=6cm, width=12cm]{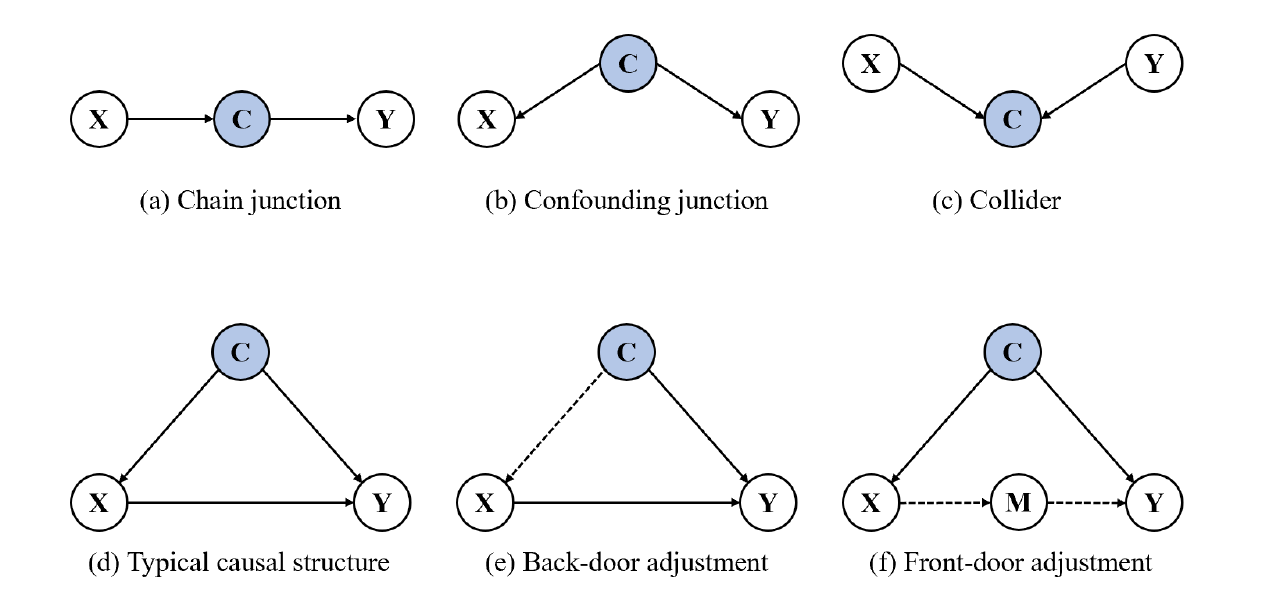}
    \caption{The three graphs (a), (b), and (c) in the first row show three common connection structures between $X$, $C$, and $Y$. They are the chain junction, the confounding junction, and the collider, respectively. Different interventions are required for different types of connections. Graph (d) in the second row is the most common causal structure, while graphs (e) and (f) are two typical causal intervention methods used for the structure in (d). (e) indicates the back-door adjustment and (f) indicates the front-door adjustment. In each graph, $X$ represents a set of input variables, $Y$ represents a set of output variables, and $C$ represents the confounder that causes the spurious correlation. The dashed line represents the block-out path and the blue nodes represent the confounder.}\label{causal interventions}
\end{figure*}

Causal intervention refers to the deliberate manipulation of one or more variables in a system to observe the effect on another variable. The intervention on variable $X$ is formulated with \emph{do}-calculus, $\emph{do}(X=x)$, which blocks the effect of the parent variables on $X$ and assigns $X$ to $x$. The causal relationships can be discovered through causal intervention with the modified graphical model and the corresponding manipulated conditional probabilities. The example in Fig. \ref{data process} can be taken for a better illustration. A causal intervention on the domain can be proposed to observe the effect of domain
variables on the corresponding labels. When keeping other variables constant, $P(Y|\emph{do}(D))$ can denote the outcome of $Y$ when changing domain knowledge.\\
Due to the complexity of the relationship between variables, it is difficult to directly determine the causal chain between the cause and the effect without being interrupted by the spurious correlation. Therefore, interventions are needed to ensure independence between variables when exploring the simple direct causal association. The basic connection structures proposed by Rebane \emph{et al.} \cite{rebane2013recovery} between three variables $X$, $Y$, and $C$ are shown in Fig. \ref{causal interventions} (a), (b), and (c). Different ways of intervention are used towards different structures owing to their unique characteristics. The path $X\xrightarrow{} C \xrightarrow{} Y$ in Fig. \ref{causal interventions} (a) is a chain junction where $X$ affects $Y$ via the mediator $C$. In vision tasks, the feature can be learned from the given image, and the label is made referring to the learned feature. It is easy to find that an intervention on $C$ can easily block the path between $X$ and $Y$. The path $X \xleftarrow{} C \xrightarrow{} Y$ in Fig. \ref{causal interventions} (b) is called a confounding junction, where $C$ affects both $X$ and $Y$ and is called a confounder. In vision tasks, the context can affect both the images and the labels, adding a spurious correlation to the actual causal chain between images and labels. Under these circumstances, interventions should be taken on $C$ to block the path. The path $X\xrightarrow{} C \xleftarrow{} Y$ in Fig. \ref{causal interventions} (c) is called a collider, where both $X$ and $Y$ decide $C$. In vision tasks, the image can be generated by both the content and domain information. When the value of $C$ is unknown, $X$ and $Y$ are independent. Once the value of $C$ is accessible, $X$ is in relation to $Y$. Therefore, the value of $C$ cannot be intervened to block the path between $X$ and $Y$.  \\
In conclusion, causal intervention guarantees the independence between variables and eliminates the effect brought by potential confounders. It is flexible and is decided by the specific structure. Two typical ways of causal interventions will be described below in detail.
\subsubsection{Back-door adjustment}
Back-door adjustment is one of the de-confounding techniques used in the causal intervention \cite{pearl2009causality}. Any path from $X$ to $Y$ that starts with an arrow pointing into $X$ is a back-door path. Assume that there are three variables, $X$, $Y$, and $C$; the corresponding structure is shown in Fig. \ref{causal interventions} (d). Since the back-door path $X\xleftarrow{} C \xrightarrow{} Y$ is the confounding junction structure, the way to block the path is to manipulate $C$ through an intervention \cite{castro2020causality} if the value of $C$ is available. The specific manipulation is to stratify $C$ and calculate the average causal effect at each stratum. The corresponding back-door adjustment is shown in Fig. \ref{causal interventions} (e), and the formulation is as follows:
\begin{equation}
P(Y|\emph{do}(X))=\sum_{c}P(Y|X,c)P(c).
\end{equation}
With the back-door adjustment, the probability of variable $Y$ conditional on a given variable $X$, which indicates the causal relationship can be obtained by calculating the sum of the conditional probabilities $P(Y|X,c)$ and the stratified probability of the confounder $C$.

\subsubsection{Front-door adjustment}
If $C$ is unavailable, the manipulation cannot be conducted and the back-door adjustment is useless. Under such circumstances, the front-door adjustment is responsible for guiding the discovery of causality \cite{pearl2009causality}. The observed mediator variable $M$ is introduced to block out the relationship between $X$ and $Y$, as shown in Fig. \ref{causal interventions} (f). In the presence of $M$, the effect of $X$ on $M$ and the effect of $M$ on $Y$ can be calculated, respectively.\\
To infer the effect of $X$ on $M$ as $P(M=m|\emph{do}(X))$, the path $X\xleftarrow{}C\xrightarrow{}Y\xrightarrow{}M$ is blocked out \cite{castro2020causality}. With the existence of a collider, the equation can be expressed as follows:
\begin{equation}
    P(M=m|\emph{do}(X))=P(M=m|X).
\end{equation}
To infer the effect of $M$ on $Y$ as $P(Y=y|\emph{do}(M))$, the path $W\xleftarrow{}X\xleftarrow{}C\xrightarrow{}Y$ should be blocked out. As $C$ is not available, $X$ should be controlled to block the path, and the corresponding equation can be expressed as below:
\begin{equation}
    P(Y=y|\emph{do}(M))=\sum_{x}P(Y|C,x)P(x).
\end{equation}
In conclusion, the front-door adjustment can be described as:
\begin{equation}
\begin{aligned}
     P(Y|\emph{do}(X))=&\sum_{m}P(M=m|\emph{do}(X))P(Y=y|\emph{do}(M))\\
    =&\sum_{m}P(M=m|X)\sum_{x}P(Y|M=m,x)P(x).
\end{aligned}
\end{equation}
With the front-door adjustment, the probability of variable $Y$ being conditional on a given variable $X$ can be obtained by introducing a mediator variable $M$. In conclusion, both the back-door adjustment and the front-door adjustment can estimate the causal effect. When choosing the de-confounding techniques, the correlations between variables should be justified, and the characteristics of the confounders should be figured out. If the confounder is available, a back-door adjustment is appropriate; otherwise, the front-door adjustment is the better choice.
\subsubsection{Counterfactual}
Counterfactual is another way of making causal inference. It is the opposite of the factual and is often used to estimate the difference between the variables of interest and their observed values in the real world \cite{gao2022causal}. Using the error terms \cite{pearl2009causality} through comparison, an intervention on the variables of interest is used to predict the outcome in the counterfactual world. For
example, some counterfactual features can be generated randomly using noise to measure the effectiveness of the learned image features on the final label determination. The error terms of the label performance can act as a criterion for effectiveness.
\subsubsection{Potential outcome models}
The potential outcome models are first proposed by Neyman\emph{et al.} \cite{splawa1990application} to estimate the causal effect
of a treatment variable on an outcome variable without requiring the causal graph. Ideally, the difference between two potential outcomes can be regarded as the causal effect of the treatment on the outcome. For example, given binary treatments $T=0/1$, the individual treatment effect (ITE) for an individual $i$ is defined as $Y_1^i-Y_0^i$ \cite{gao2022causal}. However, only one of these results can be observed at a time, either $Y_1^i$ or $Y_0^i$. As a result, the average treatment effect (ATE) is proposed as an extension for the individual treatment effect to measure the overall average. The formulation is as follows:
\begin{equation}
        {ATE}=\mathbb{E}[Y_1^i-Y_0^i]=\frac{1}{N}\sum_{i=1}^N(Y_1^i-Y_0^i),
\end{equation}
where $i=\{1,2,...,N\}$ represents each individual number in the population.
\section{Causality in different tasks}\label{section 3} 
\subsection{Methodology summary}
The process of methods based on causal theory includes building the structural causal models through causal discovery and choosing the proper way of causal inference. In the structure, nodes indicate the important variables or variables that are hidden but have an effect on the important ones, while the edges indicate inner causal relationships. Combining different causal structures and different concerns, suitable causal reasoning methods are selected to eliminate spurious correlations. Table \ref{table all} summarizes the causality-based methods both in vision and vision-language tasks. In addition to the task names, the causal structures and the concerns are also listed to compare different methods.\\
\subsubsection{Typical causal structures}
\begin{figure*}[t]
    \centering
    \includegraphics[height=6cm, width=12cm]{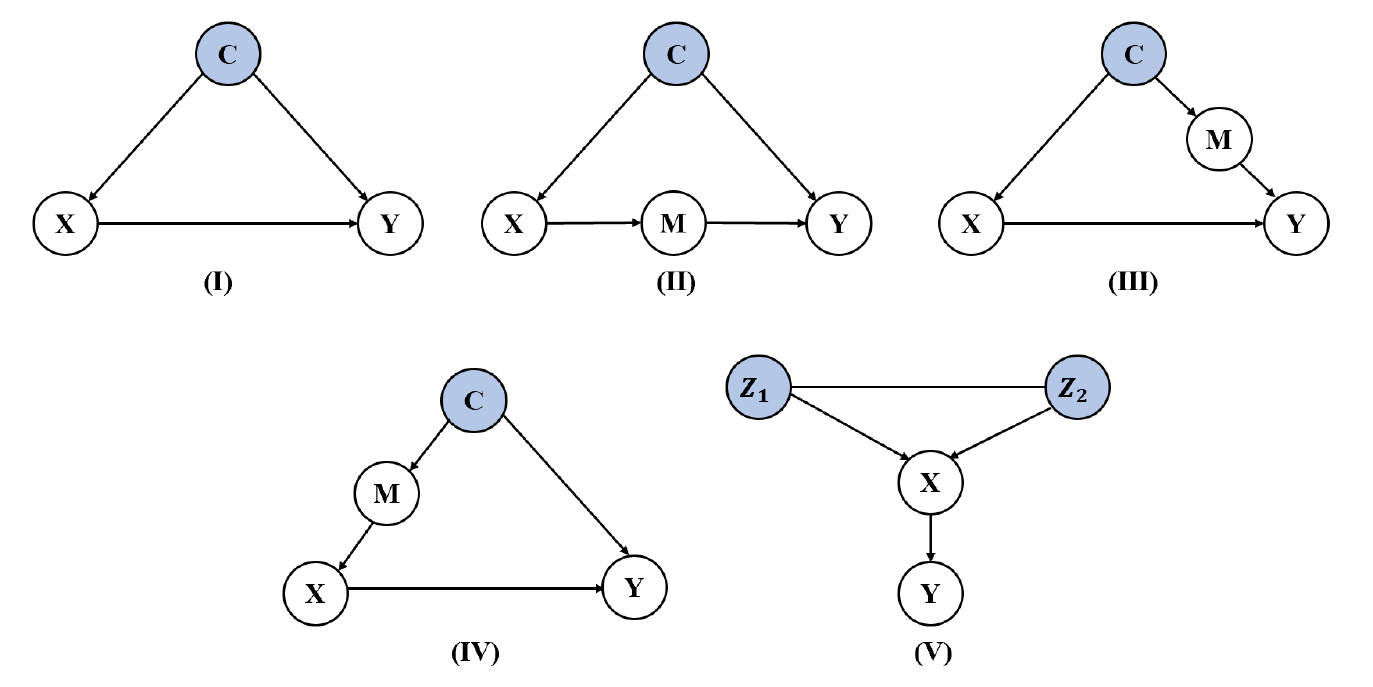}
    \caption{These are the common structures involved in causality-based tasks, in which each colored part indicates confounders.}\label{equdiagram}
\end{figure*}

The causal structures covered in this paper can be summarized into five ones, as shown in Fig. \ref{equdiagram}. 
Apart from the simplest structure (\uppercase\expandafter{\romannumeral1}) in Fig. \ref{equdiagram}, the other structures introduce intermediate variables to further elucidate the relationship. \\
Fig. \ref{equdiagram} (\uppercase\expandafter{\romannumeral1}) describes the relationships between the input, the output, and a confounder that introduces the unwanted spurious correlation. If $C$ is available, a back-door adjustment is often used to cut off the link between $X$ and $C$ to analyze the actual effect of $X$ on $Y$ as follows: 
\begin{equation}
        P(Y|\emph{do}(X))=\sum_{c}P(Y|X,c)P(c).
\end{equation}
Fig. \ref{equdiagram} (\uppercase\expandafter{\romannumeral2}) introduces an intermediate variable $M$ between the input and the output. In vision and vision-language tasks, $M$ is a more specific representation of the mapping between inputs and outputs. It is also introduced to avoid the manipulation of $C$ when it cannot be obtained directly in the front-door adjustment.
\begin{equation}
    P(Y|\emph{do}(X))=\sum_{m}P(M=m|X)\sum_{x}P(Y|M=m,x)P(x).\label{front-door}
\end{equation}
Fig. \ref{equdiagram} (\uppercase\expandafter{\romannumeral3}) introduces a new variable $M$ between the confounder and the output, which often indicates a high-dimensional feature of the effect of the confounder on the output. If the confounder $C$ is available, the intervention method will also change.
\begin{equation}
    P(Y|\emph{do}(X))=\sum_{c}P(Y|X,M,c)P(M|X,c)P(c).
\end{equation}
Otherwise, a mediator is needed to be introduced between $X$ and $Y$, as shown in Eq. \ref{front-door} for front-door intervention.\\
The intermediate variable $M$ can also be placed between the confounder and the input, as shown in Fig. \ref{equdiagram} (\uppercase\expandafter{\romannumeral4}). In this case, conditioning on the intermediate variable $M$ is equivalent to cutting off the spurious correlation.
\begin{equation}
    P(Y|\emph{do}(X))=\sum_{c}P(Y|X,c)P(c,M=m).
\end{equation}
Different from the previous structures, the structure in Fig. \ref{equdiagram} (\uppercase\expandafter{\romannumeral5}) focuses on the effect of different co-occurring confounders on the input. De-correlation of the co-occurring variables or elimination of the effect of these variables are desired in this situation. The potential outcome models proposed by Neyman \emph{et al.} \cite{splawa1990application} to quantitatively extrapolate results under different conditions often used.\\
\subsubsection{Methods for different concerns}
The main concerns in introducing causal theory for performance enhancement can be divided into three aspects: accuracy, generalizability and interpretability. These three aspects have their priorities but overlap with each other.\\
When enhancing model interpretability from a causal perspective, the approach focuses on the intrinsic causal relationship between variables and develops further inference through this objective stable correlation. Methods that use causal theory for accuracy enhancement \cite{tang2020long,hu2021distilling,huang2021causal} often focus on the task in one given domain and rebuild the structure involving the hidden confounders. They analyze the limitations of existing statistical correlation-based methods at a holistic level and identify the causes of the confounders. With the structural causal model built on prior knowledge, such methods strive to cut off the spurious correlation and inference with the right causal chain. In these methods, confounders are often the specific variables in the inference process of the task. Methods that aim at improving generalizability \cite{shen2017image,liu2021learning} often focus on the differences of one task between multiple domains and aim to achieve stable results against the domain gap. The confounders in these methods are often the domain and the corresponding context information. For generalizability, such methods aim to eliminate the effects of the domain and learn the domain-invariant features for subsequent inference. Methods that focus on interpretability \cite{goyal2019explaining,chen2020counterfactual} often expect the model to rely on the correct visual region when making decisions. Instead of making interventions under the guidance of causal structure, such methods often use the potential outcome models and the counterfactual to estimate the impact of the variables as the basis for determining network behavior. Generative models are often used in these methods to eliminate variables with no causal relationship.\\
When enhancing generalizability from the causal perspective, the spurious correlation brought by the context can be eliminated. In this case, accuracy can also be enhanced since the noise in the context will affect the exploration of causal relationships even in a domain. Guided by prior knowledge, the structural causal model makes the method inherently interpretable.\\
These three concerns are used to delineate the existing articles in the following section as methods for accuracy enhancement, methods to improve generalizability, and methods to promote interpretability. Since existing work does not fully encompass these three concerns, some tasks will lack concerns. Nevertheless, the application of causal theory for performance enhancement is meaningful in every task.
\begin{table*}[!t]
    \footnotesize
    \centering
    \renewcommand{\arraystretch}{1.4}
    \caption{This is a summary of causal theory-based vision tasks (vision-language included). For convenience, abbreviations are used to denote individual tasks: $CLS$ for classification, $DET$ for object detection, $Sem Seg$ for semantic segmentation, $WSSS$ for weakly supervised semantic segmentation, $Med Seg$ for medical image semantic segmentation, $Vis Recog$ for target recognition, $Img Cap$ for image captioning, $VQA$ for visual question answering, $3D Recon$ for 3D reconstruction, $3D Pose$ for 3D pose estimation, and $Obj Nav$ for object navigation. The structures in the last column correspond to the structures in Fig. \ref{equdiagram}.}
    \label{table all}
    \begin{tabular}{ccccccc}
    \hline
    \makebox[0.01\textwidth]{Paper}&\makebox[0.01\textwidth]{Task}&\makebox[0.01\textwidth]{Year}&\makebox[0.01\textwidth]{Concerns}&\makebox[0.03\textwidth]{Causal Instrument}&\makebox[0.02\textwidth]{Confounder}&\makebox[0.05\textwidth]{Structure}\\
    \hline
    \cite{shen2017image}&$CLS$&2017&generalizability&Causal Inference&Context&(\uppercase\expandafter{\romannumeral1})\\
    \cite{goyal2019explaining}&$CLS$&2019&Interpretability&Potential outcome models&Confounded concept&(\uppercase\expandafter{\romannumeral5})\\
    \cite{tang2020long}&$CLS$&2020&Accuracy&Back-door adjustment&Momentum in the optimizer&(\uppercase\expandafter{\romannumeral3})\\
    \cite{yue2020interventional}&$CLS$&2020&Accuracy&Back-door adjustment&Pre-trained knowledge&(\uppercase\expandafter{\romannumeral3})\\
    \cite{hu2021distilling}&$CLS$&2021&generalizability&Potential outcome models&Old knowledge&(\uppercase\expandafter{\romannumeral1})\\
    \cite{mahajan2021domain}&$CLS$&2021&generalizability&Causal inference&Domain&(\uppercase\expandafter{\romannumeral1})\\
    \cite{liu2021learning}&$CLS$&2021&generalizability&Causal inference&Context&(\uppercase\expandafter{\romannumeral1})\\
    \cite{sun2021recovering}&$CLS$&2021&generalizability&Causal inference&Dataset bias&(\uppercase\expandafter{\romannumeral1})\\
    \cite{yue2021transporting}&$CLS$&2021&generalizability&Causal inference&Unobserved semantic feature&(\uppercase\expandafter{\romannumeral1})\\
    \cite{miao2022domain}&$CLS$&2022&generalizability&Front-door adjustment&Domain&(\uppercase\expandafter{\romannumeral2})\\
    \cite{lv2022causality}&$CLS$&2022&generalizability&Causal inference&Causal factors&(\uppercase\expandafter{\romannumeral1})\\
    \cite{wang2022causal1}&$CLS$&2022&generalizability&Causal inference&Unobserved latent variable&(\uppercase\expandafter{\romannumeral1})\\
    \cite{wang2022contrastive}&$CLS$&2022&generalizability&Potential outcome models&Domain&(\uppercase\expandafter{\romannumeral1})\\
    \cite{yang2023treatment}&$CLS$&2023&Accuracy&Potential outcome models&Unobservable variable&(\uppercase\expandafter{\romannumeral1})\\
    \cite{qiu2023cafeboost}&$CLS$&2023&Accuracy&Back-door adjustment&Task identifier&(\uppercase\expandafter{\romannumeral1})\\
    \cite{chen2023meta}&$CLS$&2023&generalizability&Counterfactual&Semantic concept&(\uppercase\expandafter{\romannumeral1})\\
    \cite{huang2021causal}&$DET$&2021&Accuracy&Back-door adjustment&Context&(\uppercase\expandafter{\romannumeral3})\\
    \cite{resnick2021causal}&$DET$&2021&generalizability&Causal inference&Scene&(\uppercase\expandafter{\romannumeral1})\\
    \cite{lin2022causal}&$DET$&2022&Accuracy&Back-door adjustment&Contrast distribution&(\uppercase\expandafter{\romannumeral2})\\
    \cite{li2022disentangle}&$DET$&2022&Accuracy&Back-door adjustment&Knowledge of pre-trained model&(\uppercase\expandafter{\romannumeral3})\\
    \cite{xu2023multi}&$DET$&2023&generalizability&Causal inference&Non-causal factors&(\uppercase\expandafter{\romannumeral5})\\
    \cite{shen2021conterfactual}&$Sem Seg$&2021&Accuracy&Counterfactual&Real feature&(\uppercase\expandafter{\romannumeral1})\\
    \cite{li2022causal}&$Sem Seg$&2022&Accuracy&Back-door adjustment&Context&(\uppercase\expandafter{\romannumeral1})\\
    \cite{zhang2020causal}&$WSSS$&2020&Accuracy&Back-door adjustment&Context&(\uppercase\expandafter{\romannumeral3})\\
    \cite{wang2022causal}&$WSSS$&2022&Accuracy&Front-door adjustment&Class-specific characteristics&(\uppercase\expandafter{\romannumeral2})\\
    \cite{chen2022c}&$WSSS$&2022&Accuracy&Causal inference&Context&(\uppercase\expandafter{\romannumeral1})\\
    \cite{ding2022carts}&$Med Seg$&2022&generalizability&Causal inference&Domain shift&(\uppercase\expandafter{\romannumeral5})\\
    \cite{ouyang2022causality}&$Med Seg$&2022&generalizability&Causal inference&Acquisition process&(\uppercase\expandafter{\romannumeral3})\\
    \cite{qin2021causal}&$Vis Recog$&2021&Accuracy&Back-door adjustment&Dataset bias&(\uppercase\expandafter{\romannumeral4})\\
    \cite{liu2022contextual}&$Vis Recog$&2022&Accuracy&Back-door adjustment &Context&(\uppercase\expandafter{\romannumeral3})\\
    \cite{wang2021causal}&$Vis Recog$&2021&generalizability&Back-door adjustment&Context&(\uppercase\expandafter{\romannumeral2})\\
    \cite{mao2021generative}&$Vis Recog$&2021&generalizability&Causal inference&Unobserved confounder&(\uppercase\expandafter{\romannumeral4})\\
    \cite{mao2022causal}&$Vis Recog$&2022&generalizability&Back-door adjustment&Concept features&(\uppercase\expandafter{\romannumeral3})\\
    \cite{yang2021deconfounded}&$Img Cap$&2021&Accuracy& Front-door adjustment&Pre-training dataset&(\uppercase\expandafter{\romannumeral1})\&(\uppercase\expandafter{\romannumeral2})\\
    \cite{chen2021dependent}&$Img Cap$&2021&Accuracy&Causal inference&Context&(\uppercase\expandafter{\romannumeral2})\\
    \cite{liu2022show}&$Img Cap$&2022&Accuracy&Back-door adjustment&Visual and linguistic features&(\uppercase\expandafter{\romannumeral2})\&(\uppercase\expandafter{\romannumeral4})\\
    \cite{niu2021counterfactual}&$VQA$&2021&Accuracy&Counterfactual&Language bias&(\uppercase\expandafter{\romannumeral1})\\
    \cite{agarwal2020towards}&$VQA$&2020&generalizability&Counterfactual&Spurious correlation and bias&(\uppercase\expandafter{\romannumeral1})\\
    \cite{zhang2020devlbert}&$VQA$&2020&generalizability&Back-door adjustment&Dataset biases&(\uppercase\expandafter{\romannumeral1})\\

    \cite{chen2020counterfactual}&$VQA$&2020&Interpretability&Counterfactual&Language bias&(\uppercase\expandafter{\romannumeral1})\\
    \cite{li2022invariant}&$VQA$&2022&generalizability&Causal inference&Causal scene&(\uppercase\expandafter{\romannumeral5})\\
    \cite{zang2023discovering}&$VQA$&2023&Accuracy&Causal inference&Domain&(\uppercase\expandafter{\romannumeral5})\\
    \cite{vqaliu}&$VQA$&2023&Accuracy&Back-door and Front-door adjustment&unobserved confounder&(\uppercase\expandafter{\romannumeral4)}\\
    \cite{liu2022structural}&$3D Recon$&2022&Accuracy&Causal inference&Image&(\uppercase\expandafter{\romannumeral1})\\
    \cite{zhang2021learning}&$3D Pose$&2021&generalizability&Causal inference&Content&(\uppercase\expandafter{\romannumeral1})\\
    \cite{zhang2023layout}&$Obj Nav$&2023&Accuracy&Potential outcome models&Observation&(\uppercase\expandafter{\romannumeral1})\\
    \hline
    \end{tabular}
\end{table*}
\subsection{Causal reasoning in vision tasks}
This section will cover four tasks: classification, detection, segmentation, and visual recognition. The methods for each task will be classified according to the different concerns: accuracy, generalizability, and interpretability. The causal structure for analysis and the causal inference methods will also be discussed. The approaches that focus on accuracy often construct causal structures, analyze confounders in specific problems, and design effective causal inference methods based on the proposed structure. The approaches that focus on generalizability aim to achieve stable performance between different domains. They learn stable causal representations through methods such as causal inference and reweighting. The approaches that focus on interpretability try to explore the reasons for network decisions by generating new samples to identify the basis for network decisions.
\subsubsection{Classification}
Classification is a fundamental problem in computer vision, which tries to analyze the correlation between the images and the corresponding labels. Existing methods \cite{chen2021crossvit,li2022mvitv2,scicls} mainly focus on the statistical correlations between the features and the labels, which will be heavily influenced by spurious correlations in the presence of noise \cite{yang2023treatment} or inconsistent distributions \cite{wang2022contrastive}. Therefore, it is important to introduce causal theory into the classification task to learn the causal relationships between the images and the labels.\\
\textbf{Methods for accuracy enhancement:} The long-tail effect unbalances the impact of different classes on momentum and seriously affects classification accuracy. Tang \emph{et al.} \cite{tang2020long} assign the SGD momentum as the confounder and use the back-door adjustment on the structure, including the feature, the SGD momentum, the projection head, and the predicted label in Fig. \ref{equdiagram} (\uppercase\expandafter{\romannumeral3}). Such \emph{do}-operator removes the “bad” confounder bias while keeping the “good” mediator bias, eliminating the negative effect of SGD momentum. It is a paradox that a stronger pre-trained model may enlarge the dissimilarity between the support set and the query set in few-shot learning, thus affecting the classification accuracy. Yue \emph{et al.} \cite{yue2020interventional} point out that the pre-trained knowledge is the confounder that causes spurious correlations between the sample features and class labels in the support set. They introduce the transformed representation of sample features and use the back-door adjustment in the structure in Fig. \ref{equdiagram} (\uppercase\expandafter{\romannumeral3}) to cut off the effect of pre-trained knowledge on the feature representation. The model's misperception of the imposed noise in the image affects the classification accuracy. Qiu \emph{et al.} \cite{qiu2023cafeboost} find the new type of bias as task-induced bias and use the back-door adjustment to the structure in Fig. \ref{equdiagram} (\uppercase\expandafter{\romannumeral1}) between image, label and task identifier to transforms biased features into unbiased ones. Yang \emph{et al.} \cite{yang2023treatment} summarized the effects of noise as unobserved confounders. Guided by the structure in Fig. \ref{equdiagram} (\uppercase\expandafter{\romannumeral1}), they use the generative model to generate unobserved confounders for estimation and assess causal effects in noisy image classification tasks through treatment estimation. The learning of the robust representation is proposed against any unexpected noise.\\
\textbf{Methods to improve generalizability:} In practical image classification tasks, the assumption of independent identical distributions is unrealistic. In an ever-changing environment, there exist stable variables that remain constant and unstable variables that often change. The instabilities, in turn, create biases between domains, introduce spurious variables, and affect generalizability. Concerning the need for multi-domain generalization, methods are proposed with different structures. Lv \emph{et al.} \cite{lv2022causality} build a structural causal model containing causal factors, non-causal factors, raw inputs, and the category label based on Fig. \ref{equdiagram} (\uppercase\expandafter{\romannumeral1}). They use a causal intervention module, a causal factorization module, and an adversarial mask module for a more robust representation learning. With the same structure, a causal regularizer is proposed by Shen \emph{et al.} \cite{shen2017image} to balance the confounder distributions for each treatment feature by reweighting the samples. Through reweighting, the confounder distributions can be balanced to correct for bias from the non-random treatment assignments. By adopting a different structure, as shown in Fig. \ref{equdiagram} (\uppercase\expandafter{\romannumeral2}), Miao \emph{et al.} \cite{miao2022domain} use the front-door adjustment due to the specific causal structure between seven variables: the domain, the object, the prior knowledge, the category factor, the domain identity, the unseen images, and the labels. They transfer unseen images to taught knowledge which are the features of seen images and cut off excess causal paths to calculate the causal effect. Chen \emph{et al.} \cite{chen2023meta} propose a new learning paradigm, namely simulate-analyze-reduce to infer the causes of domain shift between the auxiliary and source domains during training. Through counterfactual inference, they reduce the effect of domain shift between semantic concept, image and category in Fig. \ref{equdiagram} (\uppercase\expandafter{\romannumeral1}).\\
\textbf{Methods to promote interpretability:} It is essential to explain the classification decision drivers of the neural network, however, existing correlation-based explanation methods fail to consider the impact of the confounders and are easily affected by misleading information. Goyal \emph{et al.} \cite{goyal2019explaining} propose the conditional generative model for the generation of counterfactuals and quantitatively measure the concept by causal concept effect method. With the structure shown in Fig. \ref{equdiagram} (\uppercase\expandafter{\romannumeral5}), they model the relationships between the high-level concepts, the images, and the classifier output.\\
\subsubsection{Detection}
The goal of object detection is to determine where objects are located in a given image and to which category each object belongs. Different from the classification task, the detection task requires not only an accurate classification but also the precise location of the target in the image. Existing methods \cite{zeng2022small,tecdet,scidet} have made significant achievements in practical applications concerning autonomous driving, but their effectiveness is heavily compromised by biases from complex scenarios. The introduction of causal theory into the detection tasks allows for accurate categorization by learning stable target properties on the one hand and precise localization by sorting out the causal interactions between the target and the environment on the other.\\
\textbf{Methods for accuracy enhancement:} Existing object detection methods tend to focus on the statistical correlation between instances, bounding boxes, and labels, ignoring spurious correlations introduced by contextual bias, which in turn results in decreased accuracy. Huang \emph{et al.} \cite{huang2021causal} formulate the causalities in object detection tasks with the structure in Fig. \ref{equdiagram} (\uppercase\expandafter{\romannumeral3}) and assign the context as the confounder. Through back-door adjustment, the non-causal but positive correlation between pixels and labels can be avoided. In unsupervised salient object detection tasks, biases caused by semantic contrast and image distribution heavily introduce spurious bias, thus limiting the improvement of performance in accuracy. To eliminate both the contrast distribution bias and the spatial distribution bias, Lin \emph{et al.} \cite{lin2022causal} identify the visual contrast distribution as the confounder which misleads the model training towards data-rich visual contrast clusters. They use the causal inference in the structure shown in Fig. \ref{equdiagram} (\uppercase\expandafter{\romannumeral4}) to make each visual contrast cluster contribute fairly and propose an image-level weighting strategy to eliminate the spatial bias. The application of knowledge distillation improves the model’s ability to capture semantic information in few-shot object detection. However, the empirical error of the teacher model degenerates the student model’s prediction of the target labels, interfering with the student’s prediction accuracy on downstream tasks. Li \emph{et al.} \cite{li2022disentangle} designate classification knowledge for specific domains as the confounder and use the back-door adjustment in the structure in Fig. \ref{equdiagram} (\uppercase\expandafter{\romannumeral4}) to remove the effect of the semantic knowledge and re-merge the remaining variables as new knowledge distillation targets.\\
\textbf{Methods to improve generalizability:} Automated driving technology requires extremely high levels of safety and robustness assurance. Perception modules trained on neural networks are often used for target detection. However, modules with better training results often fail to guarantee performance in unknown scenarios due to deviations between the target and training domains. Resnick \emph{et al.} \cite{resnick2021causal} collect real-world data through an automated driving simulator and make causal interventions on the data to discriminate factors detrimental to safe driving. In turn, the source of the domain variation problem is resolved in advance, and the deviation is eliminated. Xu \emph{et al.} \cite{xu2023multi} propose to remove non-causal factors from common features by multi-view adversarial training on source domains to remove the non-causal factors and purify the domain-invariant features. They clearly clarify the relationships among causal factors, noncausal factors,
domain specific feature and domain common feature from a causal perspective.\\
\subsubsection{Segmentation}
\begin{figure*}[t]
    \centering
    \includegraphics[height=6cm, width=15cm]{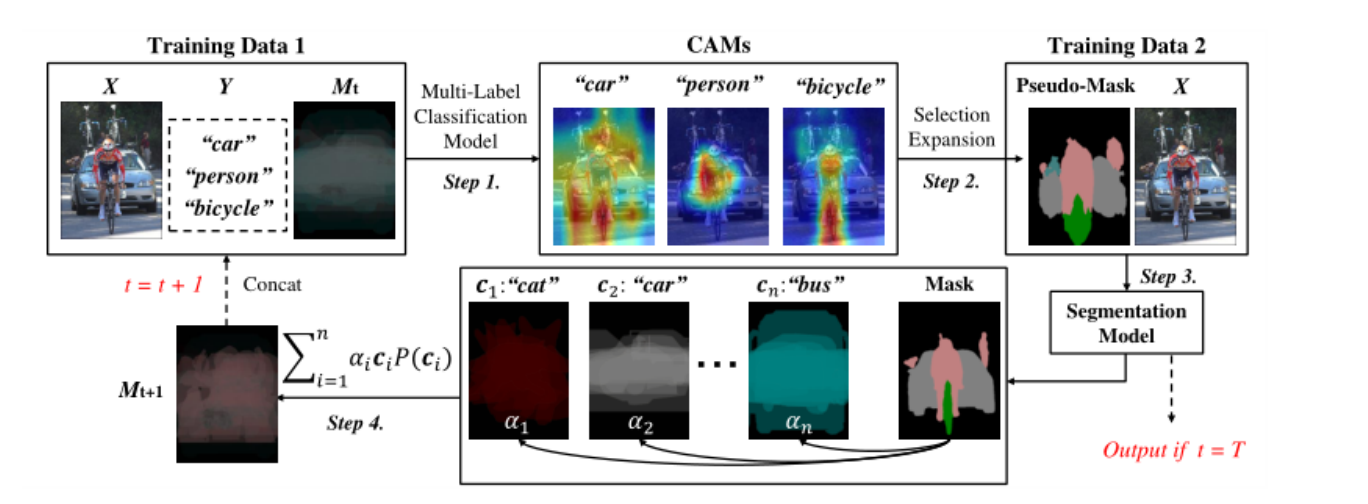}
    \caption{This is the overview of the proposed Context Adjustment (CONTA) in \cite{zhang2020causal}. In their setting, $Y$ represents a set of given labels indicating the specific classes of the objects, $c_{i}$ represents the average mask of the i{\textrm{th}} class images, and $M$ can be viewed as a linear combination of $c_{i}$. This weakly
supervised semantic segmentation task aims to enhance the quality of the generated class activation maps.} Since the confounder proposed is unobserved, it uses an iterative procedure to establish $P(c)$ in a back-door adjustment. The confounders are better learned through iteration, resulting in high-quality class activation maps for segmentation.
    \label{CONTA}
\end{figure*}

Segmentation is equivalent to a pixel-wise classification problem as the most demanding task in terms of accuracy, and the spatial-semantic uncertainty principle is the main challenge \cite{geng2018survey}. Among the causality-based segmentation tasks, fully supervised and weakly supervised segmentation problems tend to focus on accuracy, while medical segmentation problems are more concerned with generalizability to ensure safety.\\
\textbf{Methods for accuracy enhancement:}
The wrongly explored semantic correlations between word embeddings and visual features in generative models can lead to spurious correlations and compromise performance. To eliminate the bias between visible and invisible classes caused by confounders in zero-sample semantic segmentation, Shen \emph{et al.} \cite{shen2021conterfactual} adopt the counterfactual theory with the causal structure containing true and false features, word embeddings, and labels. With the same structure as Fig. \ref{equdiagram} (\uppercase\expandafter{\romannumeral1}), a causal structure consisting of images, category labels, and contextual information is constructed by Li \emph{et al.} \cite{li2022causal}. The confusion bias is eliminated through causal intervention, while a fusion module is designed to fuse original features and causal contextual relationships for a closer resemblance to the human learning process. For high-quality seed regions in weakly supervised semantic segmentation, Zhang \emph{et al.} \cite{zhang2020causal} construct the structural causal model between pixels, contexts, and labels as shown in Fig. \ref{equdiagram} (\uppercase\expandafter{\romannumeral3}) and introduce a back-door adjustment to eliminate the negative impact of contexts on class activation graph generation as shown in Fig. \ref{CONTA}. 
Different from the settings in \cite{zhang2020causal}, Wang \emph{et al.} \cite{wang2022causal} attribute the poor performance of existing methods to a set of class-specific latent confounders in the dataset and analyze the causality between image, image-level tag, pixel-level localization, and a set of class-specific latent confounders, referring to Fig. \ref{equdiagram} (\uppercase\expandafter{\romannumeral2}). They use the front-door adjustment to cut off the spurious correlation between the confounder and the images. Focusing on weakly supervised semantic segmentation in medical images, an approach that focuses on both the category-causality chain and the anatomy-causality chain is proposed by Chen \emph{et al.} \cite{chen2022c}. They use causal interventions to get the actual cause of category prediction and integrate anatomical constraints for the actual cause of segmentation.\\
\textbf{Methods to improve generalizability:} 
Generalizability is a top priority in medical image segmentation due to the personal safety involved. Causal reasoning is frequently used in medical image processing to mitigate domain shifts caused by imaging modalities, scanning protocols, and device manufacturers \cite{ding2022carts,ouyang2022causality}. A causal structure involving the operating system, the surgical environment, and the segmentation graph is built by Ding \emph{et al.} \cite{ding2022carts}. They align tool models with image observations by updating the initially incorrect robot kinematic parameters through forward kinematics and differentiable rendering to optimize im-
age feature similarity end-to-end. With a more formal causal structure between acquisition, content, feature maps, medical images, and the ground-truth segmentation mask, Ouyang \emph{et al.} \cite{ouyang2022causality} propose a simple causality-inspired data augmentation approach to expose a segmentation model to synthesized domain-shifted training examples.\\
\subsubsection{Visual recognition}
Visual recognition aims to imitate the human visual system as much as possible. Previous works \cite{yuan2022volo,srinivas2021bottleneck,scirec} have noticed the inconsistency between the training set and the application environment, compensating for the bias effect through data augmentation, re-weighted loss, and normalization. However, such methods cannot design generic methods of eradicating deviations for different scenarios. As a result, the revisiting of the fundamental process of building a visual recognition system is processed with the help of causal theory to distinguish the paradoxical character of the bias.\\
\textbf{Methods for accuracy enhancement:} Contextual bias misdirects attention to the co-occurrence context rather than the objects, leading to the loss of accuracy.  Liu \emph{et al.} \cite{liu2022contextual} build the structure between the object representations, prior context knowledge, image-specific context, and predictions. They propose a novel paradigm with both back-door adjustment and counterfactual inference to conquer the effect of contextual bias with the structure in Fig. \ref{equdiagram} (\uppercase\expandafter{\romannumeral3}). Dataset bias always contributes to the biased statistic correlation-based model, resulting in decreased performance. Qin \emph{et al.} \cite{qin2021causal} argue that such bias misleads the correlation between the input images and the output labels. They build the causal structure between image, label, context, common sense, and bias, referring to the structure in Fig. \ref{equdiagram} (\uppercase\expandafter{\romannumeral4}), and cut off the backdoor path involving the confounder by the back-door adjustment.\\
\textbf{Methods to improve generalizability:}  Since the association between images and labels is not generalizable across domains, the out-of-distribution performance is always poor. Guided by the causal-transportability language \cite{bareinboim2013general}, Mao \emph{et al.} \cite{mao2022causal} build a causal structure between the input images, the corresponding labels, and unobserved variables encoding external sources of variation not captured in the images and the labels themselves with the structure in Fig. \ref{equdiagram} (\uppercase\expandafter{\romannumeral1}). From a more complex structure in Fig. \ref{equdiagram} (\uppercase\expandafter{\romannumeral2}), Mao \emph{et al.} \cite{mao2021generative} build a causal structure between six variables. Variables include object features, unobserved confounders, background features, images, and labels. Aware that attention mechanisms can no longer be robustly characterized in any confusing environment, Wang \emph{et al.} \cite{wang2021causal} focus on the deficiency of the attention mechanism when generating robust representations and propose a causal attention module that self-annotates the context confounders in an unsupervised fashion.\\
\subsection{Causal reasoning in vision-language tasks}
This section will cover two tasks: image captioning and visual question answering. The concerns, corresponding causal structure, and causal inference methods will also be discussed. Different from vision tasks, vision-language tasks aim to combine the information from vision, and language to perform complex tasks that mimic human behavior \cite{du2022survey,scivl}. The main training steps for visual-language tasks involve encoding images and texts into single-modal embeddings for representation learning, designing an encoder to integrate information from both modalities, and using different aggregation methods. Despite the great success of existing methods \cite{marino2019ok,zhou2020unified}, the bias between modalities cannot be ignored, as the semantics of images and language differ significantly and are all susceptible to the influence of the environment. These confounding factors confuse the model regarding the causal chains between important information and thus deteriorate performance. In order to avoid bias, it is necessary to reconsider the link between two modalities from a causal viewpoint in addition to learning stable characteristics from each modality separately.
\subsubsection{Image captioning}
Image captioning is a task that aims to decipher automatically the semantic information contained in an image and produce an accurate description of it \cite{scicap}. Numerous efforts \cite{cornia2020meshed,pan2020x} have been made to improve the performance of image captioning systems, but the endogenous language bias is neglected since the existing image captioning models are inclined to build spurious connections between images and the high-frequency concurrent categories. By applying causal theory to this task, it is necessary to learn a stable representation from the images and analyze the causal relationships between images and text for appropriate responses.\\
\textbf{Methods for accuracy enhancement:} Since the dataset bias is inevitable, the accuracy performance is always affected by spurious associations caused by the bias. Yang \emph{et al.} \cite{yang2021deconfounded} introduce the semantic structure set as a mediator and use the front-door adjustment in Fig. \ref{equdiagram} (\uppercase\expandafter{\romannumeral2}) combined with the back-door adjustment in Fig. \ref{equdiagram} (\uppercase\expandafter{\romannumeral1}) to eliminate the spurious correlation caused by language resources, which is denoted as the confounder. As a result, the dataset bias introduced by the confounder is reduced. Noticing that Yang \emph{et al.} \cite{yang2021deconfounded} neglect the confounded visual features in the encoder, Chen \emph{et al.} \cite{chen2021dependent} eliminate the spurious correlation between visual features and certain expressions by using the back-door adjustment and estimate the confounder with variational inference. Considering the limitation in \cite{chen2021dependent} that the pre-training dataset is hard to stratify, the confounder is explicitly divided into two classes by Liu \emph{et al.} \cite{liu2022show}.
In most transformer-based image captioning methods, the bias caused by both visual and linguistic confounders is often overlooked. However, the resulting spurious correlations often compromise the accuracy of the network. Liu \emph{et al.} \cite{liu2022show} propose to disentangle the region-based features and use the back-door adjustment to deconfound the causal structure as shown in Fig. \ref{equdiagram} (\uppercase\expandafter{\romannumeral2}) and (\uppercase\expandafter{\romannumeral5}), which can effectively eliminate the spurious correlations caused by both visual and linguistic confounders.
\subsubsection{Visual question answering}
\begin{figure*}[t]
    \centering
    \includegraphics[height=6cm, width=10cm]{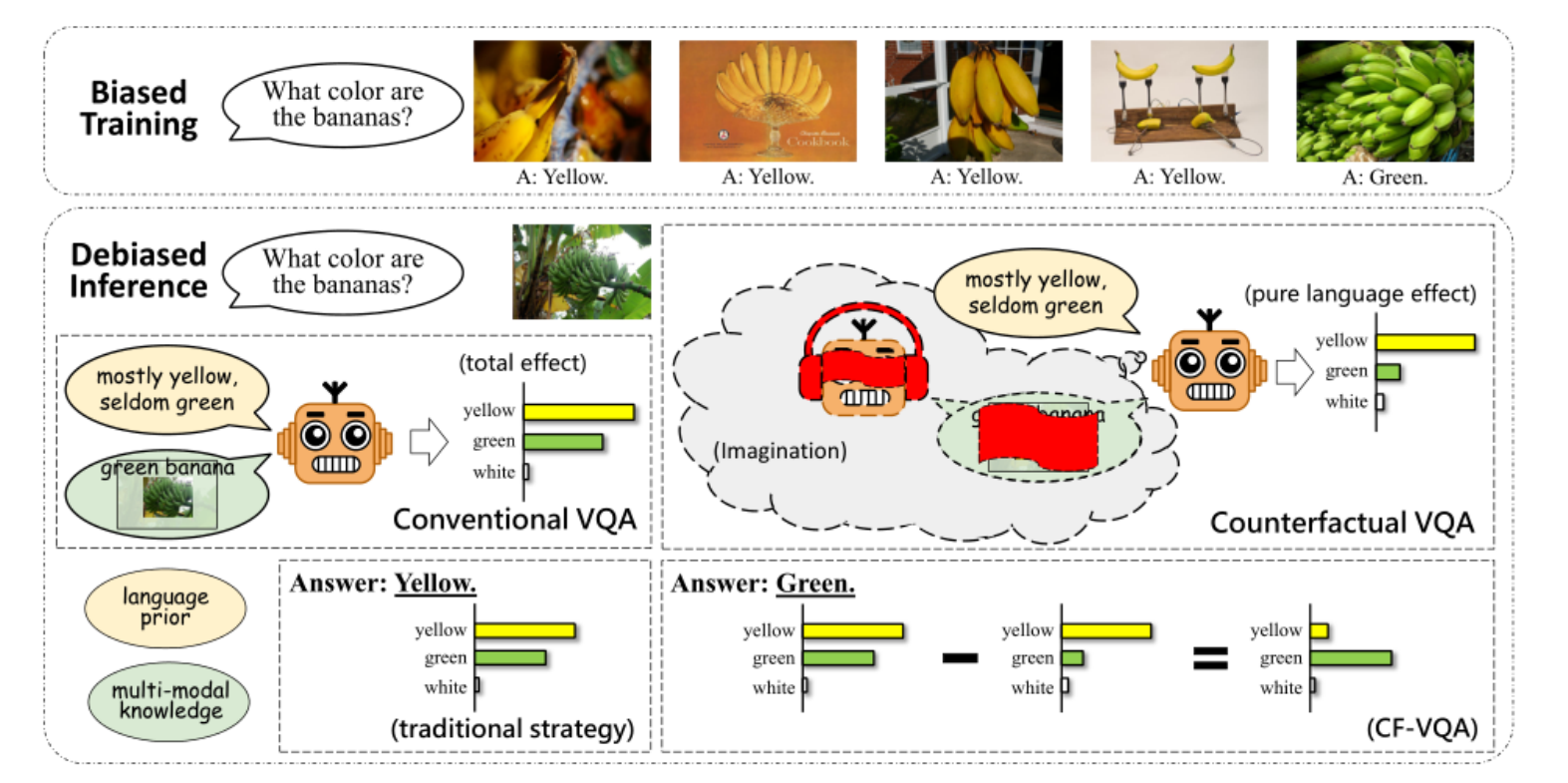}
    \caption{The biased training section shows the negative impact of language bias. The trained model will focus more on the question than the visual content, which seriously limits the generalization. The debiased strategy is proposed for a better generalization when the distributions are inconsistent across different stages. With this motivation, a counterfactual inference framework is constructed based on language-prior knowledge in \cite{niu2021counterfactual}.}
    \label{vqa NIU}
\end{figure*}

Given an image-question pair, visual question answering (VQA) tasks expect the model to answer questions correctly based on the given information. Since the existing models are proven to be fragile to linguistic variations in questions and answers \cite{ray2019sunny}, the causal relationships between images and language information are investigated to grasp the meaning of the inquiry.\\
\textbf{Methods for accuracy enhancement:}
Due to the inclusion of both image and language modal data, VQA requires visual analysis, language understanding, and multi-modal reasoning. However, existing models suffer from linguistic bias and do not make correct use of semantic information. Niu \emph{et al.} \cite{niu2021counterfactual} refer to Fig. \ref{equdiagram} (\uppercase\expandafter{\romannumeral1}) to analyze this problem and mitigate the language bias by subtracting the direct language effect from the total causal effect with the counterfactual inference as shown in Fig. \ref{vqa NIU}. Zang \emph{et al.} \cite{zang2023discovering} are concerned about the bias introduced by multi-modal data in VQA tasks. They capture visual features related to the semantics of the question and weaken the influence of local language semantics on the question answering. Liu \emph{et al.} \cite{vqaliu} collaboratively disentangle the spurious correlations between vision and language modal through back-door adjustment and front-door adjustment. They capture the fine-grained interactions between visual and linguistic semantics and learn the global semantic-aware visual-linguistic representations adaptively.
\\
\textbf{Methods to improve generalizability:} 
Recent VQA models are proven to be brittle to linguistic and semantic variations in questions and images. Agarwal \emph{et al.} \cite{agarwal2020towards} propose to enhance the robustness through automated semantic image manipulations and tests for consistency in model predictions. A data augmentation strategy that involves adding noise to the input image to improve model performance on challenging counting questions is also introduced. Li \emph{et al.} \cite{li2022invariant} propose a learning framework to ground the question-critical scene for invariant inference across different scenes.\\
\textbf{Methods to promote interpretability:} Enhancing the interpretable aspects of the model helps to understand the basis on which the model is used in decision-making. However, existing methods are guided by attention and this statistic-based approach is vulnerable to language bias. Also, there are issues with existing methods such as the need for additional annotations. Chen \emph{et al.} \cite{chen2020counterfactual} consider the VQA task as a multi-class classification problem and use counterfactual analysis to synthesize the samples, encouraging the model to perceive the difference in questions when changing some critical words and clarify the basis for their decisions.\\
\section{Future roadmaps}\label{section 4}
\subsection{Reasonable causal structure}
Though causality has shown great potential in improving accuracy, generalizability, and interpretability, all the approaches are premised on the causal structures that are tailored to the specific problems \cite{liu2022causal}. However, many of the variables involved in the vision tasks are simply ignored and only some important ones are treated as nodes to construct the simplified structure. Different choices of variables can lead to different structures, thus leading to different perspectives on the same task. When solving the problem of weakly supervised semantic segmentation, Zhang \emph{et al.} \cite{zhang2020causal} construct the causal structure between pixels, contexts, and labels, while Chen \emph{et al.} \cite{chen2022c} build the structure between images, categories, context confounders, anatomical structures, shapes of segmentation, and pseudo masks. Although addressing the same issues, the two approaches have their focus, one on contextual information and the other on biased category information. While constructing causal structures is key to causal learning, true causality is complex and there is no uniform modeling framework. When analyzing complex vision problems, it is of great importance to model a reasonable, and comprehensive causal structure for a clearer interpretation of the relationships between visually relevant variables.
\subsection{More guidance for multi-modal fusion}
Multi-modal fusion integrates different modal information to overcome the restrictions of incomplete information given by a single modality, hence realizing modal information complementarity and improving feature representation \cite{multimodal}. However, data fusion is challenging. Data generation is driven by numerous underlying processes that depend on many inaccessible variables, while the data itself is always heterogeneous and complex. Data fusion aims to allow modalities to communicate entirely and inform each other. As a result, selecting an analytical model that properly depicts the link between modalities and provides a meaningful combination of them is critical. \cite{multimodalieee}. Causality can uncover causal chains between variables objectively and can further guide analytical model building between multi-modal data to enhance accuracy, generalizability, and interpretability. For example, image data enables the precise visualization of the location of problems, and time-series data offers a thorough history of changes in the pertinent variables. A reliable causal chain is constructed by detecting faults with image data and analyzing the root cause of faults with time-series data, enabling more accurate fault detection and root cause identification. The interpretability of causal theories can also enhance the interactive ability of multi-modal fusion.
\subsection{More applications of causal theory}
So far, the applications of causal theory to vision and vision-language tasks have been discussed. Since the causal perspective can explore the relationships between variables at a fundamental level, it is suitable for guiding the study of complex systems and tasks, such as industrial processes \cite{rob,robb}, electrical systems \cite{powers}, and Earth systems \cite{cli}. Robots must make stable and timely decisions in complex industrial processes in response to the ever-changing environment. It is also important to make the decision-making process of robots transparent to humans. Recently, the task plan explanation methods \cite{lindner2022step,daruna2022explainable} have promoted human-robot interaction in robotics and enhanced the interpretability of robotic decision-making mechanisms with the help of causality. The generalization techniques also make the deployment of tasks in unknown environments a reality. The power system is a unified whole capable of producing, transmitting, distributing, and consuming electrical energy. Recently, with the improvement of top-level design, the power system is developing towards safety, high efficiency, low carbon, and intelligent integration. However, the strong interactions between systems introduce new challenges in maintaining high supply security, as new factors can affect the overall security of the power system \cite{YOHANANDHAN2022107720}. The causal theory can be used to explain the relationships between power system variables, respond to power system trends, and assist researchers by providing them with more interpretable analysis. In large-scale complex dynamical systems such as the Earth system, anthropogenic intervention is uncontrolled and against humanitarianism. To find out the mechanisms between such complex systems, the causal theory is the best choice \cite{runge2019inferring}. With an explosion in the availability of large-scale time series data and an increasingly accurate perception of the Earth model, the causal interdependencies of the underlying system can be discovered and quantified to improve the theoretical understanding of the Earth system. Combining time series, satellite remote sensing images, and other multi-source data, the nonlinear dynamical interactions between different climate factors can be analyzed and the effective prevention of tipping points can be predicted. In addition, a deeper understanding of the influence of meteorological parameters on meteorological phenomena can be obtained.
\section{Conclusion}\label{section 5}
This review aims to contribute to the development of causality in typical computer vision tasks and provide detailed explanations of existing methods from the perspective of causal structures. The concept and necessity of causality in guiding vision tasks are first discussed. The survey on existing methods of causal reasoning in vision and vision-language tasks is then conducted from different perspectives: accuracy, generalizability, and interpretability. Finally, future roadmaps are suggested to encourage the theory and applications of this promising field.

\bibliography{main}

\end{document}